# E-CONAN (Entailment, CONtradition And Neutral) Diagnostics Dataset
# Investigating Linguistic Phenomena in Arabic Natural Language Understanding

Khloud AL Jallad[1*], Nada Ghneim[2], Ghaida Rebdawi[1]
Khloud.aljallad@hiast.edu.sy, n-ghneim@aiu.edu.sy, ghaida.rebdawi@hiast.edu.sy
[1] Informatics Department, Higher Institute for Applied Sciences and Technology, Damascus, Syria.
[2] Faculty of Informatics and Communication, Arab International University, Daraa, Syria.

***Corresponding Author**: Khloud AL Jallad

## Abstract

Natural Language Understanding (NLU) plays a crucial role in various applications, yet its performance suffers from weaknesses in handling the complexities of human languages, ranging from lexical ambiguity to high-level reasoning difficulties. Analyzing errors across diverse linguistic phenomena is crucial for NLU improvement, as it will help humans get insights to comprehensively assess models' limitations and capabilities, so optimizing models' generalization. Notably, several benchmarks contain diagnostics datasets designed for investigation and fine-grained error analysis across a wide range of linguistic phenomena. When highlighting the gaps in the state-of-the-art, we noted that there is no naming convention for macro and micro categories or even a standard set of linguistic phenomena that should be covered. To overcome this gap, we propose an initial hierarchy for Cross-Lingual NLU error analysis. Moreover, we propose a methodology to create an NLI hierarchical framework and applied a case study on Arabic NLU. This framework can be applied to evaluate and investigate the performance across diverse wide range of linguistic phenomena in diverse NLU tasks, making the overall performance evaluation less data dependent. Moreover, this paper introduces E-CONAN Diagnostics dataset, a freely available dataset manually-annotated with coarse-grained and fine-grained categories based on our proposed Arabic hierarchy. E-CONAN dataset helps NLU designers better understand their models by doing error analysis and in-depth investigation. We used E-CONAN to investigate the performance of 9 pretrained language models and 5 LLMs. Results indicate that while the LLMs generally achieved almost equal results with pretrained baselines, LLMs outperform pretrained models in world knowledge and commonsense reasoning macro-category, and underperform pretrained models in syntactic macro-category. Moreover, we note that the hardest phenomena for all pretrained models and LLMs is *Reasoning* macro-category. Also, we can note that the easiest phenomena for all pretrained models is *Syntactic*, and the easiest for LLMs is *Lexico-Syntactic*.

## 1. Introduction

Natural Language Understanding (NLU) is essential for many applications, but its effectiveness is hindered by the complicated nature of human languages. NLU systems have limitations as struggling with natural languages issues ranging from ambiguous simple lexical level to complex reasoning tasks. Getting a high score on NLU tasks benchmarks does not provide sufficient insights to comprehensively assess a model's limitations or capabilities. Therefore, there is a need to have a detailed structured data for important linguistic phenomena that may cause specific limitations and weaknesses. Optimizing NLU models to overcome those weaknesses requires addressing them through error analysis across diverse linguistic phenomena and in-depth investigation using Diagnostics datasets.

Diagnostics dataset is not a test set for machine learning models performance. It is a specialized evaluation dataset that is used by humans to pinpoint specific areas where models struggle. Diagnostics dataset allows in-depth investigation through error analysis across diverse linguistic phenomena. By analyzing model performance on these specific phenomena, NLU designers can better understand their model's generalization behavior, gain valuable insights and develop targeted strategies to overcome NLU weaknesses. This could ultimately lead to more robust and generalizable models.

As for the structure of Diagnostics data, each instance is a pair of sentences tagged with the coarse-grained categories corresponding to the linguistic phenomena. Each of these coarse-grained categories has several fine-grained subcategories.

Inspired by [1], our Diagnostics dataset (E-CONAN), has the following coarse-grained categories: Lexical, Lexico-syntactic, Syntactic, Discourse, Logic, Knowledge and Common Sense. In addition to these categories, we introduced the Pragmatics coarse-grained category. All categories with their subcategories will be discussed in detail in section 3.

The contributions of this paper are:

- **Initial Cross-Lingual Hierarchy**: Recently, several benchmarks contain diagnostics datasets were designed for NLU performance investigating and fine-grained error analysis across a wide range of linguistic phenomena. When highlighting the gaps in the state-of-the-art, we noted that there is no naming convention for macro and micro categories or even a standard set of linguistic phenomena that should be covered. To overcome this gap, we propose an initial hierarchy for Cross-Lingual NLU error analysis. This hierarchy can be used as an initial step when building a diagnostics dataset in any language. While this hierarchy covers a broad range of linguistic phenomena, we believe that there is still a room for improvement. We hope that this hierarchy will be a significant step towards a more comprehensive and effective hierarchy for all languages in NLU evaluation.
- **Arabic Language Hierarchy:** we proposed a hierarchical framework for analyzing Arabic NLU tasks phenomena. This framework can be used to assess and analyze models' performance on various linguistic phenomena across different NLU challenges, making the overall performance less data dependent
- **A Comprehensive Arabic Diagnostics Dataset:** we introduced E-CONAN Diagnostics dataset, a freely available dataset manually-annotated with coarse-grained and fine-grained categories based on our proposed Arabic hierarchy. To the best of our knowledge, this is the largest dataset compared to existing NLU Diagnostics in all languages.

The paper is organized as follows: Section 1 provides an introduction. Related works are discussed in Section 2. The proposed framework is presented in Section 3. Section 4 shows dataset statistics and comparisons with SoTA Diagnostics datasets. As for results, discussions and conclusion, they are shown in sections 5, 6 respectively.

## 2. Related Works

### 2.1. Natural Language Inference

To the best of our knowledge, the first research work that defines natural language inference was in 1973 at Stanford university [2], where they outline a method for using Preference Semantics to resolve ambiguous pronouns in complex sentences. Their system handles situations where understanding the pronoun requires in-depth world knowledge or requires ability to infer events. It converts all information into a standardized format and creates potential links between unknown pronouns and possible referents.

Textual Entailment (TE) was originally proposed by Dagan and Glickman in 2004 [3] as a generic paradigm for applied semantic inference, and subsequently established through the series of benchmarks known as the *PASCAL Recognizing Textual Entailment (RTE) Challenges* from 2005 up to 2011 (Bar-Haim et al., 2006; Bentivogli, Clark, et al., 2009; Bentivogli et al., 2011; Bentivogli, Magnini, et al., 2009a, 2009b; Dagan Ido and Glickman, 2006; Giampiccolo et al., 2007, 2008a; Glickman & Dagan, 2004).
Recognizing Textual Entailment (RTE), also known as Natural Language Inference (NLI), is the task of detecting whether a sentence meaning can be entailed from another sentence. The first sentence is called Text (T) or Premise (P), the second sentence is called Hypothesis (H). TE task is to determine if a reader who reads T would entail H or not. An Example is shown in Table 1.

| T | H | Relation |
|---|---|---|
| **I love reading** | I hate reading | Contradiction |
| | I hate oranges | Neutral |
| | I like learning. | Entailment |

*Table 1: Recognizing Textual Entailment Pairs Examples*

TE task encapsulates all NLU capabilities within a very simple interface: recognizing when the meaning of a text snippet is contained in the meaning of a second piece of text or not. This simple abstraction of an extremely complex problem has made TE more common and much more important because it can be also used as a component in other NLP applications, from Machine Translation to Information Extraction [13]. When first introduced, TE task was called 2-way RTE as it classified pairs into two relations (entail, not entail). The 3-way-RTE term appeared in 2008 [14], as the task of determining entailment relation between pairs of sentences introduced three relations (entail, contradict, neutral or unknown).

Several NLI and RTE datasets were constructed in many languages. Also, several models were published by many universities and companies. We will cover few works in English and Arabic only.

As for English language, most famous available datasets are: WNLI [15], MNLI [16], SNLI [17]. Also, GLUE proposed RTE dataset [18] which is a combination of RTE1
(Dagan Idoand Glickman, 2006)
, RTE2 [11], RTE3 [10], and RTE5 [9]. Some state-of-the-art models in English are: PaLM 540B[19], Vega_v2_6B[20], RoBERTa[21], SemBERT[22], XLNET[23], AlexaTM_20B[24] , BloombergGPT and GPT-NeoX and BLOOM_176B [25], UnitedSynT5[26], ByT5[27] , Rethinking Coupling[28], mGPT[29], DeBERTa[30], SpanBERT[31] , SqueezeBERT[32] ,DistilBERT[33].

As for Arabic language, some available datasets are: ArbTEDS corpus [34], ArNLI dataset [35], ArEntil Dataset [36]. Furthermore, some multilingual datasets contain Arabic section such as SNLI dataset [37], XNLI dataset [38].

Some state-of-the-art models that contain Arabic language are: Facebook Bart Large MNLI [39], [40], Facebook RoBERTa-Large-MNLI [41] , Microsoft Multilingual MiniLM [42], Microsoft XLM-RoBERTa [42], Microsoft DeBERTaV3 [43]. Moreover, Laurer et al. tuned many versions of pretrained multilingual models such as, mDeBERTa-V3-Base-XNLI-Multilingual-NLI-2mil7 [44],Moritzlaurer Ernie-M-Base-MNLI-XNLI [44], Moritzlaurer Multilingual-MiniLMv2-L6-MNLI-XNL [44], Moritzlaurer Multilingual-MiniLMv2-L12-MNLI-XNLI[44]. We have used many of them as zero-shot classification to test and validate our proposed diagnostics dataset.

### 2.2. Linguistic Phenomena

FraCaS (Framework for Computational Semantics) is the first *linguistic phenomena hierarchy framework* for Diagnostics evaluation using computational semantics [45]. It was proposed in 1996 at Stanford university. FraCas linguistic phenomena were: generalized quantifiers, negation, monotonicity, anaphora, ellipsis, comparatives, adjectives, temporal relations, and propositional attitudes. The FraCaS corpus consists of 346 problems, each including 1-5 statements (premises), one yes/no question, and a yes (entail) / no (contradiction) / don't know (neutral) answer.
However, FraCas project did not include any evaluations, and there were no follow-ups till 2009, when MacCartney[46] conducted some corrections on FraCas corpus, added relevant notes, and rephrased the questions into declarative sentences.[1]

While FraCas systematically covered a wide range of linguistic phenomena, the small size of the dataset limited the meaningful generalization evaluation.
Later, in 2010 Bentivogli et. al [1] proposed a methodology for TE specialized dataset creation, which was not a Diagnostics dataset. Authors carried out a feasibility study applying the methodology to a sample of 90 pairs extracted from the RTE-5 dataset [4]. The studied phenomena were as following: (See Table 2)

- **lexical:** identity, format, acronymy, demonymy, synonymy, semantic opposition, hyperonymy, geographical knowledge.
- **lexical-syntactic:** transparent heads, nominalization/verbalization, causative, paraphrase.
- **syntactic:** negation, modifier, argument realization, apposition, list, coordination, active/passive alternation.
- **discourse:** coreference, apposition, zero anaphora, ellipsis, statements.
- **reasoning:** apposition, modifiers, genitive, relative clause, elliptic expressions, meronymy, metonymy, membership/representativeness, reasoning on quantities, temporal and spatial reasoning, all the general inferences using background knowledge.

Several papers analyzed TE specialized data, such as [47], and several papers were done to design an evaluation specialized dataset such as quantitative reasoning [48], comprehensive reading [49], common sense reasoning [50]. Some papers were done to create a test-suite for specific tasks such as negation in Not another Negation Benvhmark NanNLI [51], translation representation analysis [52]

The first *Diagnostics dataset* was introduced in GLUE benchmark [32]. It contained 1104 pairs, with 4 Coarse-grained Categories: Lexical Semantics, Predicate-Argument Structure, Logic, Knowledge and Common Sense. Each of them had its detailed fine-grained subcategories (See Table 2).

The first *Arabic Diagnostics dataset* was proposed in ALUE benchmark [33]. It contained 1147 pairs, with 5 Coarse-grained Categories: Lexical Semantics, Predicate-Argument Structure, Logic, Quantification, Knowledge and Common Sense. Each of them has its detailed fine-grained subcategories (See Table 2).

[1] The original version of FraCaS corpus as ps-file and its improved XML-version are freely available for download from (https://nlp.stanford.edu/~wcmac/downloads/).

| DIAGNOSTICS NAME | Coarse-grained CATEGORIES | FINE GRAINED CATEGORIES |
|---|---|---|
| **Specialized TE Dataset** | **LEXICAL** | Identity/Mismatch; Format; Acronymy; Demonymy; Synonymy; Semantic Opposition; Hypernymy; Geographical Knowledge |
| | **LEXICAL-SYNTACTIC** | Transparent Head; Nominalization/Verbalization; Causative; Paraphrase; |
| | **SYNTACTIC** | Negation; Modifier; Argument Realization; Apposition; List; Coordination; Active/Passive Alternation; |
| | **DISCOURSE** | Coreference; Apposition; Anaphora Zero; Ellipsis; Statements; |
| | **REASONING** | Apposition; Modifier; Genitive; Relative Clause; Elliptic Expression; Meronymy; Metonymy; Membership/Representative; Quantity; Temporal; Spatial; Common Background/ General Inferences |
| **GLUE Diagnostics Dataset** | **LEXICAL SEMANTICS** | Lexical Entailment; Morphological Negation; Factivity; Symmetry/Collectivity; Redundancy; Named Entities; Quantifiers |
| | **PREDICATE-ARGUMENT STRUCTURE** | Syntactic Ambiguity; Prepositional Phrases; Core Arguments; Alternations; Ellipsis/Implicits; Anaphora/Coreference; Intersectivity; Restrictivity |
| | **LOGIC** | Propositional Structure; Quantification; Monotonicity; Richer Logical Structure |
| | **KNOWLEDGE AND COMMON SENSE** | World Knowledge; Common Sense |
| **ALUE Diagnostics Dataset** | **LEXICAL SEMANTICS** | Lexical Entailment; Morphological Negation; Symmetry/Collectivity; Redundancy; Named entities; Quantifiers |
| | **PREDICATE-ARGUMENT STRUCTURE** | Relative clauses; Prepositional Phrase; **Alternation (**Causative/Inchoative; Active/Passive; Nominalization; Datives; Topicalization**);** Anaphora/Coreference; Intersectivity; Restrictivity |
| | **LOGIC** | Negation; Double negation; Conjunction; Disjunction; Conditionals; **Monotonicity (**Upward entailment; Downward entailment; non-Monotone**)** |
| | **QUANTIFICATION** | Universal; Existential |
| | **KNOWLEDGE AND COMMON SENSE** | World Knowledge; Common Sense |

*Table 2: Diagnostics Dataset Categories Comparisons*

We noticed that while existing datasets encompass a broad range of semantic phenomena, it seems that there is a need for more in-depth exploration of higher-level linguistic phenomena. This motivated us to investigate more in the pragmatic level in our proposed framework as we believe that it deserves more exploration. Moreover, unlike all previous diagnostics that were designed based on English grammars, we designed our diagnostics hierarchy upon Arabic grammar rules.

## 1. PROPOSED FRAMEWORK

In this section, we will illustrate the detailed hierarchy. Moreover, we will discuss our proposed methodologies for designing hierarchy and constructing diagnostics dataset based on hierarchy. Finally, we will discuss examples from diagnostics dataset.

### 3.1. HIERARCHY OF LINGUISTIC PHENOMENA

We grouped linguistic phenomena into categories and subcategories. Our macro (coarse-grained) categories are: Lexical, Lexico-Syntactic, Syntactic, Discourse, Logic, Knowledge & Common Sense, and Pragmatic. Each macro category includes micro (fine-grained) phenomena, as shown in Figure 1. Moreover, a case study is applied in Arabic language is shown is Figure 2 and shown in detail in Table 5 in appendix. It worth mentioning that Arabic grammar rules were presented in both Arabic and English, where the English terms for these rules were derived from an Arabic course designed for English speakers [54]. More details with Arabic and English examples are provided in Github page[2].

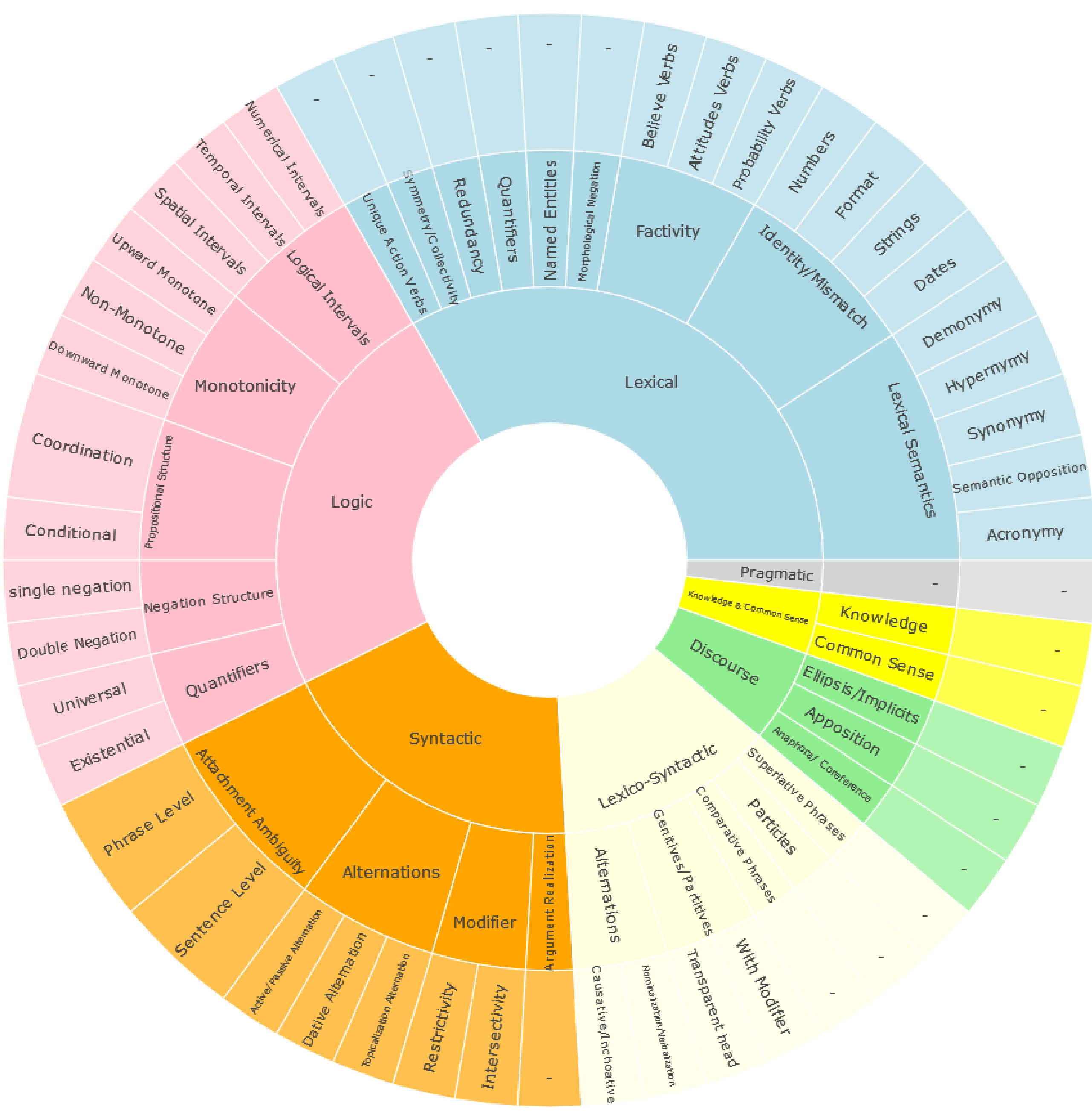

*Figure 1: Our Proposed Initial Cross-Lingual Hierarchy*

[2] Link will be provided after acceptance

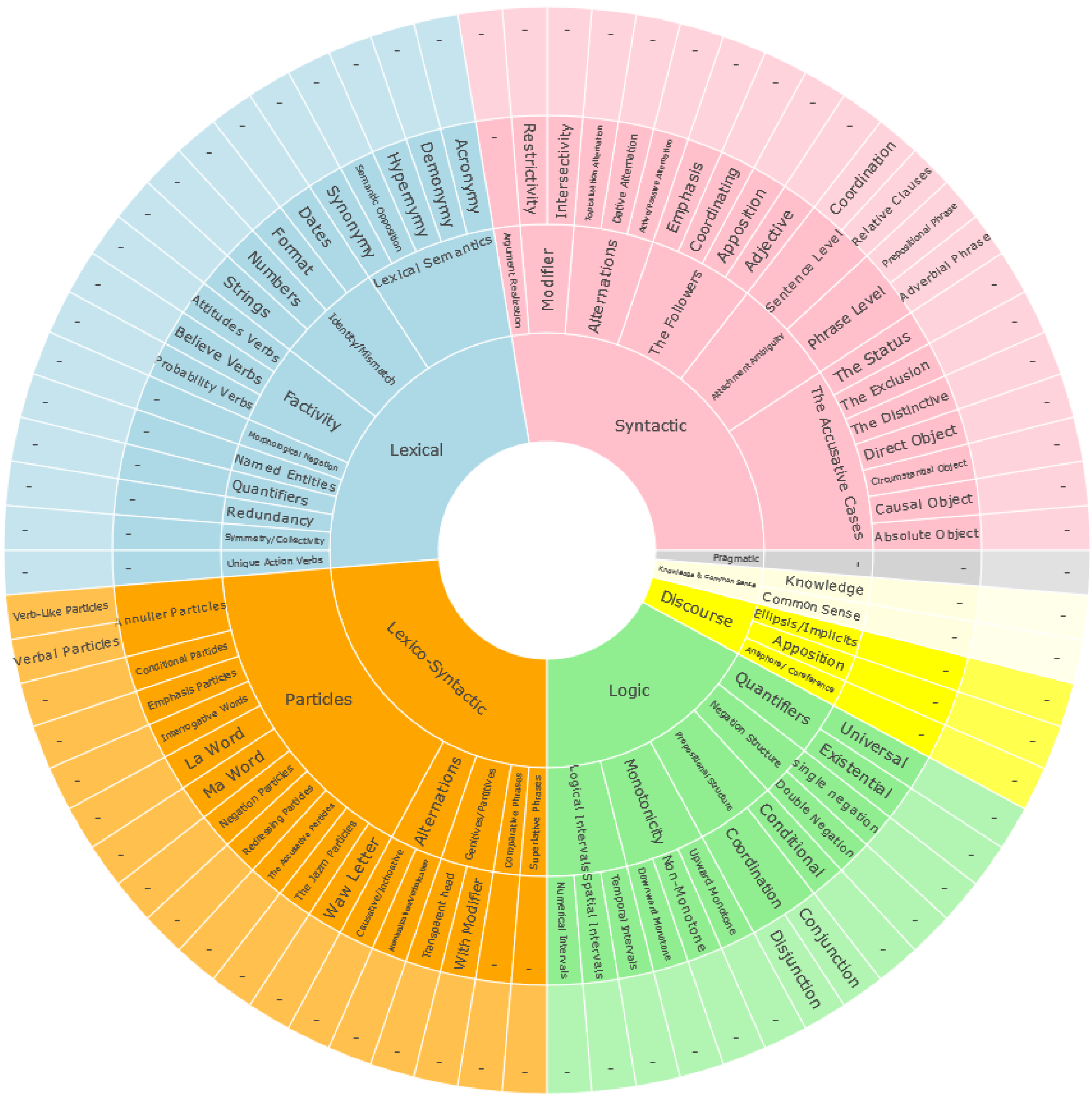

*Figure 2: Our Proposed Arabic Hierarchy*

## 3. Dataset Construction Methodology

Hierarchy was built primarily through a **bottom-up** approach, analyzing and tagging examples based on various criteria (lexical, syntactic, ...). However, a top-down approach was constrained during the acquisition of sentence groups, each containing several hundred elements. This top-down approach aimed to refine the hierarchy by grouping micro-categories into macro-categories based on Arabic grammar rules.

E-CONAN Diagnostics dataset sentences were extracted from two sources: (1) Arabic grammar books and (2) ALUE and GLUE datasets.
Regarding the creation of sentence pairs from Arabic grammar books, the following methodology was used:

1- **Data Collection:** Sentences relevant to each micro-category were extracted from Arabic grammar books such as [55], [56],[57] and from Arabic teaching websites such as Mawdoo[3][58], Loghate[4] [59] and MadinahArabic[5][54]. We make minimal modifications to each extracted sentence to maintain high lexical and structural overlap within each sentence pair and limit superficial cues. Where possible, we produce several pairs with different NLI labels for a single source sentence, to have minimal sets of sentence pairs that are lexically and structurally very similar but correspond to different entailment relationships
2- **Sentence Pair Creation:** For each extracted sentence, a second sentence was created to form a pair by changing or removing the functional word and making minimal modifications to the sentence in order to reflect different possible interpretations. Then each of them was considered once as a *premise* sentence and once as a *hypothesis* sentence.
3- **Pair Annotation:** Finally, each pair of sentences was tagged with its corresponding micro and macro categories, and the entailment relationship between sentences.

Figure 3 illustrates the methodology for sentences extracted from Arabic books. Figure 4 illustrates Arabic Sentences Pairs Creation Methodology with Example.

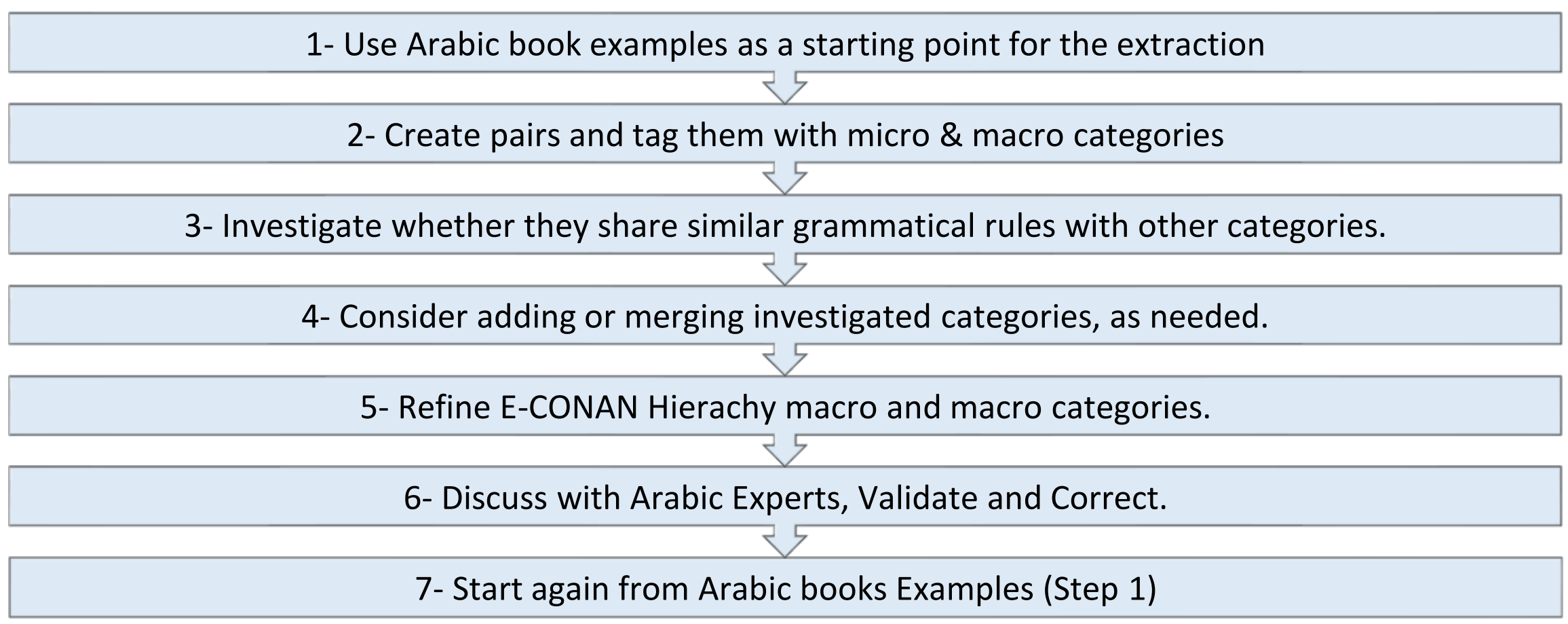

*Figure 3: Methodology for Sentences Extracted from Arabic Books.*

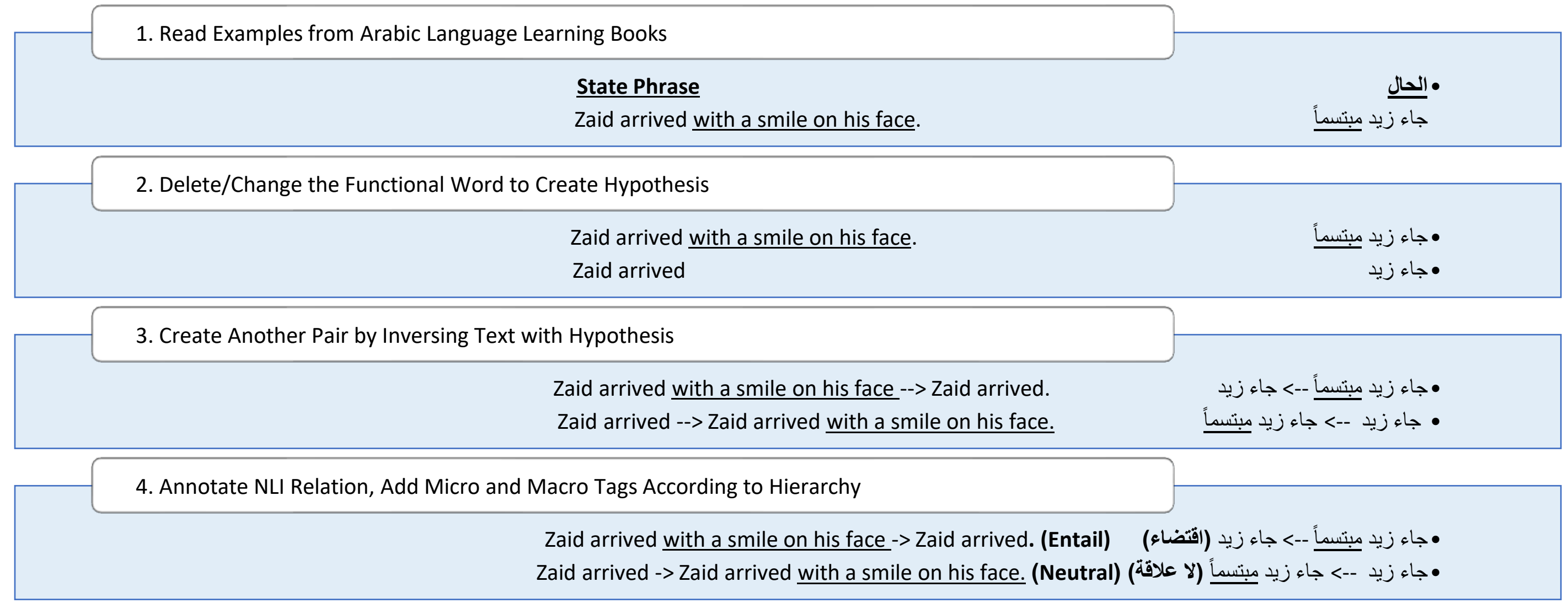

*Figure 4: Arabic Sentences Pairs Creation Methodology with Example*

[3] https://mawdoo3.com/تصنيف:قواعد_اللغة_العربية
[4] https://loghate.com/n/قواعد-النحو
[5] https://www.madinaharabic.com/arabic-language-course/lessons/

On the other hand, for pair construction based on ALUE and GLUE datasets, we used the following steps:

1. **Data Collection and Translation:** Examples were collected from both ALUE and GLUE datasets. Sentences from GLUE were then translated into Arabic.

2. **Pair Comparison and Filtering:** Comparisons were made between the corresponding ALUE and GLUE sentence pairs. Ambiguous, redundant or unclear sentences were excluded.

3. **Micro and Macro Category Tagging:** Finally, ALUE and GLUE tags were used after adapting them according to our proposed hierarchy.

As for labeling entailment relation for NLI pairs, we consider the relation based on the idea that both sentences are introduced at the same time, same context for the same entities. For instance, the sentence “there is honey inside the jar” contradicts with “there is milk inside the jar” as we are talking about the same entity “jar” at the same time. But the sentences “X rebuilt the palace” is neutral with “Y rebuilt the palace” as the rebuilt action can be done several times in the past, but at the time of speaking both sentences could be true at the same time about the same entity “palace”. Some examples from created sentences are shown and discussed in Table 6 in appendix.
E-CONAN Diagnostics dataset introduces the *Pragmatic* level as a separate macro-category. This category was not studied in previous works, as far as we know.

### 4. Statistics and Comparisons

To the best of our knowledge, E-CONAN is the largest dataset in its kind for all languages, not only for Arabic. Moreover, it includes wider macro and micro categories. Thus, it may provide deeper error-analysis with wider various examples to investigate. Although determining the necessary “amount of data required to produce relevant performance measures” remains an open problem [60].

Some statistics and comparisons of E-CONAN with SoTA are shown in Figure 5

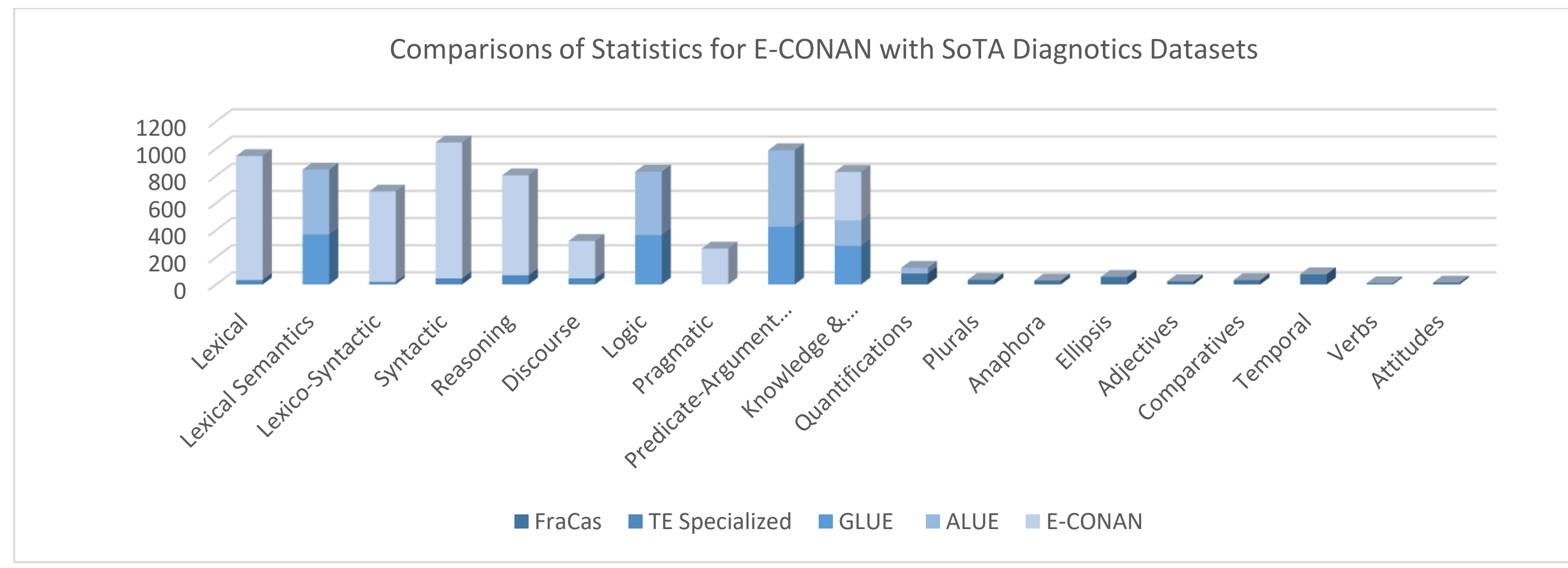

*Figure 5: Statistics E-CONAN Comparisons with SoTA*

## 4. Discussion & Analysis

While we adopted the SoTA basic macro and micro categories [61], our hierarchical structure differs. The proposed hierarchy was developed by sub-categorizing some existing categories in more detail and aggregating others to form new, more abstract levels.

In this section, we discuss the linguistics phenomena that are used in our hierarchy compared to SoTA diagnostics. Firstly, we present the definition of each phenomenon based on Arabic linguistics references. Secondly, we discuss the entailment relation labeling strategy that we designed. More detailed examples are shown in github[6].

## I. Lexical

We have created most of our samples following mixed methodology from FraCas, GLUE and TE specialized dataset as illustrated in [61].

As for *Identity/Mismatch* we enriched TE specialized with more sub categories to include the following: *Strings, Numbers, Dates, Format*.

As for *Morphological Negation*, *Redundancy*, *Named Entities*, *Quantifiers*, *and Symmetry/Collectivity* Level_1 Micro Categories and *Lexical-Semantic* sub levels (Level_2 Micro Categories): *Synonymy, Semantic Opposition (Contrast), Hypernymy*, samples were created following the same methodology in GLUE as illustrated in [18], however, we adjusted their levels within our hierarchy.

As for *Factivity* Level_1 Micro Category, it was inspired by GLUE [18], after adding three micro categories as follows:

a. Attitudes Verbs: For example, ("حادث مروري حصل اليوم", "A traffic accident happened today") has *no relation* with the sentence ("سمعت عن حادث مروري اليوم", "I heard about a traffic accident today"), while the opposite is *entailment* relation, as I may not hear about all accidents, but if I heard about an accident, it is certain that one occurred.
b. Believe Verbs: such as ("وجدت" meaning "I believe" or "I am sure that"). If a believe verb exists in one of (T, H) sentences, both sentences will entail the same meaning but with more focus in one of them than the other. For example, ("وجدت العلم نوراً", "I believe that knowledge is power.") has *entailment* relation with the sentence ("العلم نور", "Knowledge is power").
c. Probability Verbs: such as ("ظنَّ"," thought"), ("زعم", "claim"). For instance, the sentence ("ظننت الامتحان سهلاً", "I thought that the exam was easy") has *neutral* relation with the sentence ("الامتحان سهل", "the exam was easy") as not all things happen as expected.

As for *Unique Action Verbs* inspired by [62], we created our samples to include both cases: unique actions and non-unique actions. For instance, the sentence ("ولدت في سورية", "I was born in Syria") has *contradiction* relation with the sentence ("ولدت في كندا", "I was born in Canada") as the *born* verb is unique and has one unique place. In contrast, some verbs are repetitive such as the verb *travel*, thus the sentence ("سافرت إلى سورية", "I travelled to Syria") has *neutral* relation with the sentence ("سافرت إلى كندا", "I travelled to Canada") as one may travel several times during his life. Travel is not a unique action.

As for *Lexical Semantics* inspired by GLUE, we included 5 micro categories inspired by TE Specialized [1].

## II. Lexico-Syntactic

As for *Alternations* that contain *Nominalization/Verbalization*, Causative/Inchoative, we created our samples following the same methodology in GLUE as illustrated in [61], but we adjusted their level to align with our proposed hierarchy.

As for *Particles*, we studied the different categories as follows:

a) *The Arabic annuller particles*: as stated in [63], called annullers from annulment, which means shifting and suppressing. It contains (/inna/ and its sisters), (/kana/ and its sisters), and (/đhanna/ and its sisters).

First, (/inna/ and its sisters): Each of these particles adds a new meaning to the Arabic nominal sentence, as follows: /inna/ ("إنَّ", "indeed", a confirmation particle), /anna/ ("أنَّ", "that", a phrasal particle), /ka'anna/ ("كأنَّ", "as if", a similitude particle), /lākinna/ ("لكنَّ", "but", a restriction particle), /layta/ ("ليت", "if only", a wishing particle), /laξalla/ ("لعلَّ", "may be", a wish particle).

Concerning their effect on entailment relations between (T, H) sentences pair, if the particle (/inna/, /anna/, /ka'anna/) exists in one of sentences, both sentences (T, H) will entail the same meaning but with more focus in one of them than the other. On the other hand, /layta/ particle means hoping something impossible, so having it in only one of the sentences (T or H) means a *contradiction* relation. Furthermore, /laξalla/ particle means hoping something that may be possible in the future, i.e., may happen or not. For that reason, having /laξalla/ particle in T only, indicates *neutral* relation. While having /laƺalla/ particle in H only, indicates *contradiction* relation as the event has already happened in T, and we do not hope something that has already happened.

[6] Link will be provided after acceptance

*Second, (/kana/ and its sisters):* It includes particles such as: /kāna/ ("كان", "was (verb to be)"), /đhalla/ ("ظلّ", "remained in existence"), /bāta/ ("بات", "stayed the night"), /ađħā/ ("أضحى", "to bring to light"), /aŝbaħa/ ("أصبح"," to become in the morning"), /amsā/ ("أمسى"," to become in the evening"), /ŝâra/ ("صار", "to become").

*As for the entailment relations* between (T, H) sentences pair, it depends mainly on the tenses that are represented in both sentences, as these particles can alter the tense of an event. For example, the particle /kāna/ (كان) shifts the event's temporal reference to the past tense. In contrast, /sâra/ (صار) typically restricts the event's temporal interpretation to the present tense. Other particles, such as /đhalla/ (ظلّ), indicate an event that began in the past and continues into the present, establishing a continuous state. Thus, Determning the tense of the sentences is crucial for accurately determining entailment relationships. When the tenses in both sentences are identical, an *entailment* relation is typically observed. Similarly, there is an *entailment* relation if the T tense represents a complete temporal scope that includes a partial tense expressed in H. i.e., the premise covers a broader time frame that encompasses a narrower one that is presented in the hypothesis. However, if the partial tense is presented in H while T covers a different or broader scope, then the relation becomes *neutral*. For example, the sentence ("مازالت هند مريضة", "Hind is still sick") *entails* the sentence *(*"كانت هند مريضة", "Hind was sick") while the opposite is not true.

b) <u>*Arabic interrogatives (question words):*</u> as stated in [63], used for questions, and contains particles such as: /man/ for "who" questions, /mā/ and /mādhā/ for "what" questions, /limādhā/ for "why" questions, /matā/ for "when" questions, /ayna/ for "where" questions, /kam/ used for "how many (much)" questions, /kayfa/ for how questions, /Hal/ and /hamzaa/ for "yes/no" questions.
As for their effect on entailment relations between (T, H) sentences pair, a declarative sentence has a *neutral* relation with a question sentence.

*c)* <u>*Negation Particles:*</u> used for negating the meaning in nominal sentences and verbal sentences. As for nominal sentences, we may use particles such as /laysa/ ("ليس', "not to be") or /ma/ ("ما") or /la/ ("لا") particles. As for verbal sentences, there are various particles for various tenses. i.e., either past, present or future negative, such as: /laysa/ ("ليس'), /ma/ ("ما"), /la/ ("لا"), /lam/ ("لم"), /lan/ ("لن") particles, etc. [64] [65].
As for entailment relations, having any negation particle in one sentence of the (T, H) pair sentences <u>*in same tense*</u> indicates a *contradiction* relation. For example, the sentence ("قرأت هذا الكتاب", "I read this book") *has contradiction relation with* ("لم أقرأ هذا الكتاب", "I did not read this book"). While having any negation particle in one or both sentences of the (T, H) pair <u>*in different tenses*</u> indicates a *neutral* relation. For instance, the sentence( " لن أقرأ هذا الكتاب", "I will not read this book" *has no relation with* ("لم أقرأ هذا الكتاب", "I did not read this book").

*d)* <u>*Emphasis Particles:*</u> used for meaning confirmation. It can be used in nominal or verbal sentences or both, such as /Kad/("قد") /laKad/ ("لقد"), /inna/ ("إنَّ"), /anna/ ("أنَّ"), /WaAllah/ ("والله"), etc [66].
*As for entailment relations, having any emphasis particle in one sentence of (T, H) pair indicates an entailment relation.*

*e)* <u>*The Accusative Particles*</u> *[56]* These particles precede verb in present tense, causing it to enter the accusative mood (*manṣūb* - منصوب) and often signaling a specific nuance like purpose, negation, or future tense.
*As for accusative pronouns* effects on the entailment relations between (T, H) sentences pair, it depends on the meaning of the used particle. For example, having /lan/ "لن" accusative particle in only one sentence of the (T, H) pair indicates a contradiction relation as it means negation. When having /kai/ "كي" particle, which indicates the reason or the purpose, we have several scenarios: first, having /kai/ particle in T only, entails the H, as there is no problem with having less details in H than T. This means that we conclude some ideas from premise, but not all of them. For instance, s1: ("درست كي أنجح", "I studied to succeed") *entails s2: (*"درست", "I studied"). Second, having the /kai/ particle in H only, means neutral relation with premise. This is because we cannot conclude the reason or purpose from the verb alone in T. For example, s2 in the previous example does *not entail* s1. Third, having *entailment* relation when having /kai/ particle in one of the pair sentences and having in the second sentence another particle which has the same meaning as /kai/ such as /hatta/ "حتى".

f) <u>*The Jazm Particles*</u> *[55]* there are two present verb Ĵazm particles: Negative lam "لَم" which negates the action of the present verb in the past, and Prohibitive la النَّاهِيَةُ "لا", which is specifically for the second person pronouns, and should be differentiated from the Negative la النَّافِيَةِ "لا".
*As for its* effects on the entailment relations between (T, H) sentences pair, *having Jazm Particle in only one sentence of the (T, H) pair indicates a contradiction relation as it means negation.*

g) *Conditional Particles,* as stated in [63], the conditional sentence in Arabic is made up of three parts: conditional article, conditional verb, and the answer to the condition, which is the result depending on the condition.
As for its effects on the entailment relations between (T, H) pair, having a conditional sentence in only one of the pair sentences and having only one conditional parts in the other sentence indicates a *neutral* relation.

h) Redressing Particles are used for rectification and clarification to introduce new information that either contradicts or expands on a preceding statement, preventing the reader from drawing an incorrect conclusion. For example, the particle ("بل", /bal/) can function in two main ways either to expand a statement or to negate and replace a statement. For example, In the sentence ("لم أحب الفيلم فقط، بل الممثلين أيضاً","I didn't just like the movie; I also liked the actors"), /bal/ adds information, extending the scope of the speaker's preference from the movie to include the actors. In a different context, such as ("ما أحببت الفيلم بل المسلسل","I didn't like the movie; rather, I liked the series"), /bal/ completely negates the initial statement and replaces it with the correct information.

As for *Genitives/Partitives (Idafa), we have created samples for transparent head replacement or removing. For instance, the sentence ("أخمد رجال الإطفاء الحريق", "the firefighters suppressed the fire") entails ("أخمد الإطفاء الحريق" and "أخمد فريق الإطفاء الحريق", "the firefighting team suppressed the fire").*

As for *Comparative and Superlative Phrases,* it was inspired by FraCas [45] with some modifications in the hierarchy levels and adaptations to Arabic language. Having a comparative sentence in only one of the pair sentences (T, H) indicates a *neutral* relation. For instance, the sentence ("زيد أقصر من عمرو", "Ziad is shorter than Amr") has a *neutral* relation with ("زيد قصير", "Zaid is short") and ("عمرو قصير", "Amr is short"). While having different core arguments for the comparative particles indicates *contradiction* relation. For example, the sentence ("زيد أقصر من عمرو", "Ziad is shorter than Amr") has *contradiction* relation with ("عمرو أقصر من زيد", "Amr is shorter than Zaid"). On the other hand, using superlative phrase indicates different relation. For instance, the sentence ("زيد أقدم موظفي الشركة", "Zaid is the company's most senior employee") has *contradiction* relation with the sentence ("عمرو أقدم من زيد في الشركة", "Amr is more senior than Zaid at the company"). It is important to mention that superlative in idioms may have different meaning. For instance, the sentence ("حاتم أكرم الناس", "Hatem is the most generous person") has *no* relation with ("عمرو أكرم من حاتم", "Amr is more generous than Hatem") as it is an idiom in Arabic, meaning that someone is very generous. This case is not included in our samples, although we explored some pragmatic sentences.

## III. Syntactic

As for *The Accusative Cases,* we studied the different categories as follows**:**

a) *The absolute object,* it is mentioned after the verb in order to: confirm its meaning, show its nature, or specify the number. For that reason, we consider all forms of an absolute object -except specifying the number- as *entail* relations. In case of specifying the number, the relation depends whether the number is in T or H. Having it only in T entails H as there is no problem with having less details in H than T, while the opposite means *neutral* relation, as we cannot know how many times the event was repeated from having the event in T.

b) *Causal Object,* specifies the purpose or reason behind an action. It is a word that indicates *why* an action is performed.
As for affecting entailment relations between (T, H) pair, we have several scenarios: first, having different causal objects in T and H means *contradiction* relation. Second, having the causal object in T only, entails H, as there is no problem having less details in H than T. This means that we conclude some ideas from premise, but not all of them. Third, having the causal object in H only, means *neutral* relation with T, as we cannot conclude the reason behind the verb from the verb alone in T.

*c)* *The Exclusion* as stated in [63], used to exclude the noun placed after /illā/ (meaning except) or one of its sisters ("سوى", "عدا", etc.), from the rule of the words placed before.
As for the exclusion effects on the entailment relations between (T, H) pair, there are two main scenarios. First, a sentence that includes an exclusion contradicts the exact same sentence when the exclusion is removed. For example,( "نجح كل الطلاب إلا واحداً", "All students passed the exam except one" *contradicts* ("نجح كل الطلاب", "All students passed the exam"). Second, an entailment exists when having T sentence with an exclusion and H sentence without exclusion, but including quantifiers that represent a narrower subset. For instance, the statement ("نجح جميع الطلاب إلا واحداً", "All students passed the exam except one") *entails* ("نجح معظم الطلاب", "Most students passed the exam.") However, the reverse is not true (*neutral* relation).

*d)* *Al-Hal (Status),* unlike the objects, the status is not related to the verb; rather it is related to another noun in the verbal sentence which is called the concerned noun. The concerned noun can be the doer (/alfāξil/), or can be the direct object.
As for affecting entailment relations between (T, H) sentences pair, we have several scenarios: First, having different status in T and H means *contradiction* or *entailment* relation depending on whether two statuses have similar meanings or contrasted meanings. Second, having the status in T only, *entails* H as there is no problem with having less details in H than T. Third, having the status in H only, means *neutral* relation with T, as we cannot conclude the status of the verb from the verb alone in T.

*e)* *Al-tamyīz (The Distinctive)* is used to clear up the ambiguity in the preceding noun (the distinguished).
Its impact on entailment relations between (T, H) pair, is the same as the scenarios of *Al-Hal (Status).*

*f)* *Circumstantial Object*, also called the adverb and is called in Arabic (/ađħ đharf/) which literally means vessel or container, because it indicates the containing time or place in which the action of the verb occurs.
Its impact on entailment relations between (T, H) pair, is the same as the scenarios of *Al-Hal (Status).*

*Direct Object*, is “who” or “what” the action happens to [66].
Its impact on entailment relations between (T, H) pair, is the same as the scenarios of *Al-Hal (Status).*

As for *Followers***,** we proposed 4 micro categories as follows:

a) *Adjective:* means “a description”, and the word it’s describing is called the مَنْعُوْت (the “described word”). We can say it’s like an adjective in English except that in Arabic, it will come after the word it’s describing, and not necessarily right after it. “A big house” would be “بَيْتٌ كَبِيْرٌ” (Which is بَيْتٌ – “house”, followed by كَبَيْرٌ – “big”) [66].

As for Its impact on entailment relations between (T, H) pair, is the same as the scenarios of *Al-Hal (Status) and Restrivitiy.*

b) *Apposition (Substitution):* two words are placed side by side, both with the same syntactic function. It occurs when a word is replaced by another, and the substitute takes on the exact same grammatical state as the original word. It has many types complete substitution, substituting a part for the whole, and substituting content for the container. For that reason, we consider all complete substitution examples, as entail relations. While the rest types are entailment, neutral and contradiction according to used particles [66]. For instance, the sentence (“I memorized the Quran “حفظت القرآن”) *entails* the sentence (“I memorized the Quran, half of it”, “ حفظت القرآن نصفه”), but the opposite has *neutral* relation.

c) *Emphasis:* it means to strengthen something and in grammar there are two kinds of emphasis. First, (“ اَلْتَّوْكِيْد اللَّفْظِيْ”, “verbal emphasis”) – This is accomplished by repeating the word (using either the exact same word or a synonym for it). Emphasis can be done for a noun, verb or particle. Second, (emphasis by meaning) – A follower that removes the possibility that one is speaking forgetfully or intending something with a wider meaning than what he’s saying (“جاءَ الأَمِيْرُ “, “The prince came”) could give the listener the impression that you spoke forgetfully or that you really meant that the prince’s messenger came instead (“جاءَ الْأَمِيْرُ نَفْسُهُ “, “The prince, himself, came”) has ("نَفْسُهُ” “himself”) added to emphasize (“الأَمِيْرُ “, “the prince”). This removes any other possibility and establishes with the listener that the prince himself came[66].

As for Its impact on entailment relations between (T, H) pair, we considered examples of almost scenarios in both cases. For Example, (“I visited all of Damascus”, “زرت دمشق كلّها”) *entails* (“I visited Damascus”, “ زرت دمشق”), but the opposite is *neutral* relation.

d) *Coordinating:* This is a grammatical follower that is connected to what it’s following by putting one of the حروف العطف (connective particles) in between[66],There are two main types of coordination: Conjunction and Disjunction.
As for Its impact on entailment relations between (T, H) pair, we considered examples of almost scenarios in both cases. For instance, (“I drank juice and coffee”, “شربت عصيراً وقهوة”) *entails* (“I drank coffee”, “شربت قهوة”), but the opposite is *neutral* relation.

As for *Syntactic Semantic Attachment Ambiguity*, we proposed 2 micro categories: Sentence and Phrase Levels.

As for *Modifiers (which include* Intersectivity**,** Restrictivity*) and Alternations (which include Dative, Active/Passive, Topicalization)***,** we created our samples following the same methodology in GLUE, ALUE as illustrated in [18], [67], but we changed their level in the hierarchy.

**IV. Discourse**

As for *Apposition*, samples were created following the same methodology in TE, however, we adjusted their levels within our hierarchy.

As for *Anaphora/ Coreference, Ellipsis/Implicits,* samples were created following the same methodology in GLUE as illustrated in [18], however, we adjusted their levels within our hierarchy.

**V. Logic**

As for *Propositional Structure, Negation Structure, Quantifiers, Monotonicity,* we created our samples following the same methodology in GLUE, ALUE as illustrated in [18], [67], but we changed their level in the hierarchy.

As for *Logical Intervals, we proposed three sub categories:* temporal Intervals**,** numerical Intervals, and spatial Intervals.

**VI. World Knowledge & Common Sense:** we created our samples following the same methodology in GLUE as illustrated in [62].

**VII. Pragmatic**

This category was not previously included in diagnostics dataset before, even though aspects of pragmatics might have been implicitly captured within other categories. While we acknowledge that the current samples within the pragmatic category are limited in scope and may not encompass all relevant concepts, we believe this opens a door for future improvements. Examples were shown in Table 4, in section 3.

## 5. Results and Discussions

To measure the effectiveness of our created diagnostics dataset (E-CONAN), we used it to evaluate, and error-analyze for some famous state-of-the-art multilingual pretrained models using zero-shot classification:

- Facebook Bart Large MNLI[7] [39], [40] It is reported in results as **M1**
- Moritzlaurer MiniLM-L6-MNLI[8]: Microsoft Multilingual MiniLM [42] tuned on MNLI dataset [16] It is reported in results as **M2**
- Moritzlaurer Deberta-V3-Base-MNLI[9] Microsoft DeBERTaV3 [43] tuned on MNLI dataset [16] It is reported in results as **M3**
- Moritzlaurer mDeBERTa-V3-Base-MNLI-XNLI[10][44] a tuned version of a Microsoft mDeBERTa-v3-base [43] model which was trained on CC100 multilingual dataset [68][69] with 100 different languages, tuned on both MNLI [16] and XNLI [38] datasets. It is reported in results as **M4**
- Moritzlaurer mDeBERTa -V3-Base-XNLI-Multilingual-NLI-2mil7[11] [44]a tuned version of a Microsoft mDeBERTa-v3-base [43] model which was trained on CC100 multilingual dataset [68][69], tuned on both MNLI [16] and multilingual-NLI-26lang-2mil7 [44]datasets. It is reported in results as **M5**
- FacebookAI RoBERTa-Large-MNLI[12]: [41] which was trained on MNLI dataset [16]. It is reported in results as **M6**
- Moritzlaurer Multilingual-MiniLMv2-L6-MNLI-XNLI[13] [44]a Microsoft XLM-RoBERTa [42] model tuned on both MNLI [16] and XNLI [38] datasets. It is reported in results as **M7**
- Moritzlaurer Multilingual-MiniLMv2-L12-MNLI-XNLI[14][44] a Microsoft XLM-RoBERTa [42] model tuned on both MNLI [16] and XNLI [38] datasets. It is reported in results as **M8**
- Moritzlaurer Ernie-M-Base-MNLI-XNLI[15] [44] a tuned version of Meta's RoBERTa model, tuned on both MNLI [16] and XNLI [38] datasets. It is reported in results as **M9**

---

[7] https://huggingface.co/facebook/bart-large-mnli
[8] https://huggingface.co/MoritzLaurer/MiniLM-L6-mnli
[9] https://huggingface.co/MoritzLaurer/DeBERTa-v3-base-mnli
[10] https://huggingface.co/MoritzLaurer/mDeBERTa-v3-base-mnli-xnli
[11] https://huggingface.co/MoritzLaurer/mDeBERTa-v3-base-xnli-multilingual-nli-2mil7
[12] https://huggingface.co/FacebookAI/roberta-large-mnli
[13] https://huggingface.co/MoritzLaurer/multilingual-MiniLMv2-L6-mnli-xnli
[14] https://huggingface.co/MoritzLaurer/multilingual-MiniLMv2-L12-mnli-xnli
[15] https://huggingface.co/MoritzLaurer/ernie-m-base-mnli-xnli

Moreover, we used our created diagnostics dataset, (E-CONAN), to evaluate and analyze errors of some famous state-of-the-art multilingual LLMs using zero-shot prompts. The comparison was based on the following LLMs:

1. ALLAM [70]ALLaM stands for Arabic Large Language Model, a series of large language models to support the ecosystem of Arabic Language Technologies (ALT). ALLAM models are based on an autoregressive decoder-only architecture and are pretrained on a mixture of Arabic and English texts via vocabulary expansion.
2. Qwen2.5 [71], [72] is a comprehensive series of large language models (LLMs) developed by Alibaba Cloud, based on a Transformer-based decoder-only architecture. The series includes both open-weight dense models (0.5B to 72B parameters) and proprietary Mixture-of-Experts (MoE) variants (Qwen2.5-Turbo and Qwen2.5-Plus).
3. Gemma3 [73]: an open-weight, multimodal model developed by Google. Ranging in size from 1 to 27 billion parameters. It has 128K-token context window. It supports multilingual (140+ languages) and multimodal input (text and images).
4. Command R7B Arabic [74] 7-billion parameter, open-weights, multilingual LLM by Cohere Labs The model has been trained and evaluated for performance in Arabic and English, but its training data includes samples from other languages. Command R7B Arabic supports a context length of 128,000 tokens. It has been trained specifically for tasks such as the generation step of Retrieval Augmented Generation (RAG) in Arabic and English.
5. DeepSeek-R1 [75]: A large-scale, open-weight Mixture-of-Experts model from DeepSeek AI (with 671B total parameters and ≈37B active per query). It is optimized for complex reasoning and logical tasks (e.g., math, coding, scientific reasoning). The maximum generation length is set to 32,768 tokens. Its pipeline incorporates two RL stages aimed at discovering improved reasoning patterns and aligning with human preferences.

We will show only the coarse-grained results and the overall fine-grained results for each category. Although our diagnostics dataset contains much richer information and much specific breakdown detailed error analysis that are beyond the scope of this discussion as the full results are too extensive to include here. For instance, we will discuss here Particles' error rate in Lexico-Syntactic category through all models. But in the diagnostics dataset, we have labels for each particle separately, such as 'ليس', 'كان', 'كأنَّ', 'ليت', 'لعلَّ', 'صار', etc. Redressing Particles are also shown as a whole, but in datasets they are breakdown detailed to 'لكن', 'إلا' etc. And so on for all categories.
Macro-category Results are shown in Table 3,4. Where the lowest errors were marked with $^{-}$ sign, the highest errors were marked with the $^{+}$ sign, and the average errors for each category were shown in bold. Results indicate that while the LLMs generally achieved almost equal results with pretrained baselines, LLMs outperform pretrained models in knowledge-and commonsense reasoning macro-category, and underperform pretrained models in syntactic macro-category.
Detailed results are shown in Figures 6,7,8,9,10.

| | ***Lexical*** | ***Lexico-Syntactic*** | ***Syntactic*** | ***Discourse*** | ***Pragmatic*** | ***Reasoning*** | ***Knowledge & Common Sense*** |
|---|---|---|---|---|---|---|---|
| ***M1*** | **0.57**$^{+}$ | **0.61**$^{+}$ | 0.45 | **0.47**$^{+}$ | **0.58**$^{+}$ | 0.59 | **0.56**$^{+}$ |
| ***M2*** | 0.52 | 0.53 | **0.53**$^{+}$ | **0.47**$^{+}$ | 0.51 | **0.63**$^{+}$ | 0.53 |
| ***M3*** | 0.47 | 0.41 | 0.39 | 0.37 | 0.46 | 0.47 | 0.49 |
| ***M4*** | 0.34 | 0.29 | 0.28 | 0.28 | 0.35 | 0.36 | 0.37 |
| ***M5*** | **0.32**$^{-}$ | **0.24**$^{-}$ | **0.25**$^{-}$ | **0.26**$^{-}$ | **0.37**$^{-}$ | **0.32**$^{-}$ | **0.37**$^{-}$ |
| ***M6*** | 0.50 | 0.51 | 0.44 | 0.43 | 0.48 | 0.61 | 0.51 |
| ***M7*** | 0.45 | 0.43 | 0.36 | 0.32 | 0.36 | 0.49 | 0.47 |
| ***M8*** | 0.42 | 0.37 | 0.33 | 0.32 | 0.39 | 0.43 | 0.44 |
| ***M9*** | 0.40 | 0.34 | 0.30 | 0.31 | 0.41 | 0.42 | 0.43 |
| ***Average Errors*** | ***0.44*** | ***0.41*** | ***0.37*** | ***0.36*** | ***0.43*** | ***0.48*** | ***0.46*** |

*Table 3: Error Rates on SOTA Pretrained Models*

| | ***Lexical*** | ***Lexico-Syntactic*** | ***Syntactic*** | ***Discourse*** | ***Pragmatic*** | ***Reasoning*** | ***Knowledge & Common Sense*** |
|---|---|---|---|---|---|---|---|
| ***Allam*** | **0.42**$^{+}$ | **0.40**$^{+}$ | **0.46**$^{+}$ | 0.42 | **0.42**$^{+}$ | **0.48**$^{+}$ | **0.47**$^{+}$ |
| ***CMD RB*** | **0.42**$^{+}$ | 0.39 | **0.46**$^{+}$ | **0.43**$^{+}$ | 0.39 | 0.46 | 0.38 |
| ***DeepSeek*** | 0.41 | **0.40**$^{+}$ | 0.42 | 0.39 | 0.41 | **0.38**$^{-}$ | 0.43 |
| ***Gemma*** | **0.37**$^{-}$ | **0.33**$^{-}$ | **0.38**$^{-}$ | **0.37**$^{-}$ | **0.37**$^{-}$ | 0.42 | 0.34 |
| ***Qwen*** | 0.41 | 0.37 | 0.41 | **0.43**$^{+}$ | **0.42**$^{+}$ | 0.44 | **0.33**$^{-}$ |
| ***Average Errors*** | **0.41** | **0.38** | **0.43** | **0.41** | **0.40** | **0.44** | **0.39** |

*Table 4: Error Rates on SOTA LLMs*

As for pretrained models, the highest error rates were by Facebook multilingual models especially Bart, although they were trained on MNLI dataset. Lower error rates were by models tuned on both MNLI and XNLI. The lowest error rates were by MoritzLaurer/mDeBERTa-v3-base-xnli-multilingual-nli-2mil7 that was tuned on MNLI, XNLI, and multilingual-NLI-26lang-2mil7 datasets. The highest error rate in all categories was 0.63 and the lowest error rate in all categories was 0.24. We will discuss each category results: stating lowest and highest results. As for Lexical micro-category: the highest error rate was 0.57 by Bart and the lowest error rate was 0.32. As for Lexico-Syntactic micro-category: the highest error rate was 0.61 by Bart and the lowest error rate was 0.24. As for Syntactic: micro-category: the highest error rate was 0.53 by MiniLM-L6 and the lowest error rate was 0.25. As for Discourse: the highest error rate was 0.47 by MiniLM-L6 and Bart and the lowest error rate was 0.26. As for Pragmatic micro-category: the highest error rate was 0.58 by Bart and the lowest error rate was 0.37. As for Reasoning micro-category: the highest error rate was 0.63 by MiniLM-L6 and the lowest error rate was 0.32. As for World-Knowledge and Common-Sense micro-category: the highest error rate was 0.56 by Bart and the lowest error rate was 0.37.

As for LLMs, the highest error rates were by Allam, where the lowest error rate were by Gemma. The highest error rate in all categories was 0.48 and the lowest error rate in all categories was 0.33. We will discuss each category results: stating lowest and highest results. As for Lexical micro-category: the highest error rate was 0.42 by Allam, CMD-RB and the lowest error rate was 0.37 by Gemma. As for Lexico-Syntactic micro-category: the highest error rate was 0.40 by Allam, Deepseek and the lowest error rate was 0.33 By Gemma. As for Syntactic: micro-category, the highest error rate was 0.46 by Allam, CMD-RB and the lowest error rate was 0.38 by Gemma. As for Discourse micro-category: the highest error rate was 0.43 by CMD-RB, Qwen and the lowest error rate was 0.37 by Gemma. As for Pragmatic micro-category: the highest error rate was 0.42 by Allam and Qwen and the lowest error rate was 0.37 by Gemma. As for Reasoning micro-category: the highest error rate was 0.48 by Allam and the lowest error rate was 0.38 by Deepseek. As for World-Knowledge and Common-Sense micro-category: the highest error rate was 0.47 by Allam and the lowest error rate was 0.33 by Qwen.

We note that the hardest phenomena for all pretrained models and LLMs is *Reasoning* that contains the highest error rates in all of them. Also, we can note that the easiest phenomena for all pretrained models is *Syntactic*, and the easiest for LLMs is *Lexico-Syntactic*.

As for Lexical fine-grained results, Figure 6 shows that the highest error percentage were for the Hypernymy, Identity Mismatch and Morphological Negation. Lowest error rates were for Redundancy, Synonyms and Demonyms.

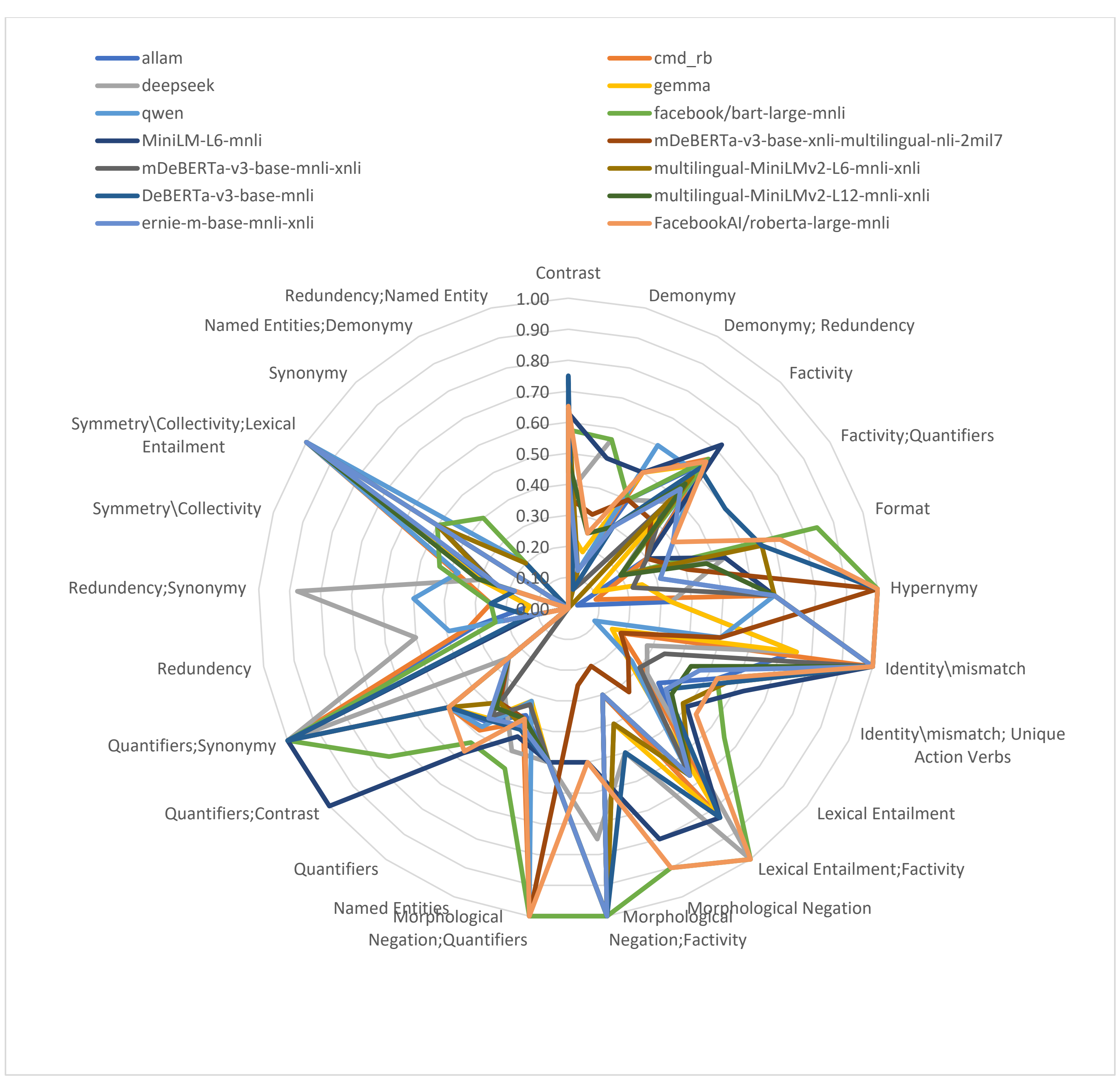

*Figure 6: Lexical Fine-grained Error Rates on E-CONAN Diagnostics Data.*

As for Common Sense & World knowledge fine-grained results, Figure 7 shows that error rates for the World knowledge are higher than Common Sense error rates. Models trained on XNLI dataset show better results in both Common Sense and world knowledge categories. Lowest error rate on all models was 30% indicating that models need to be tuned more on this category.

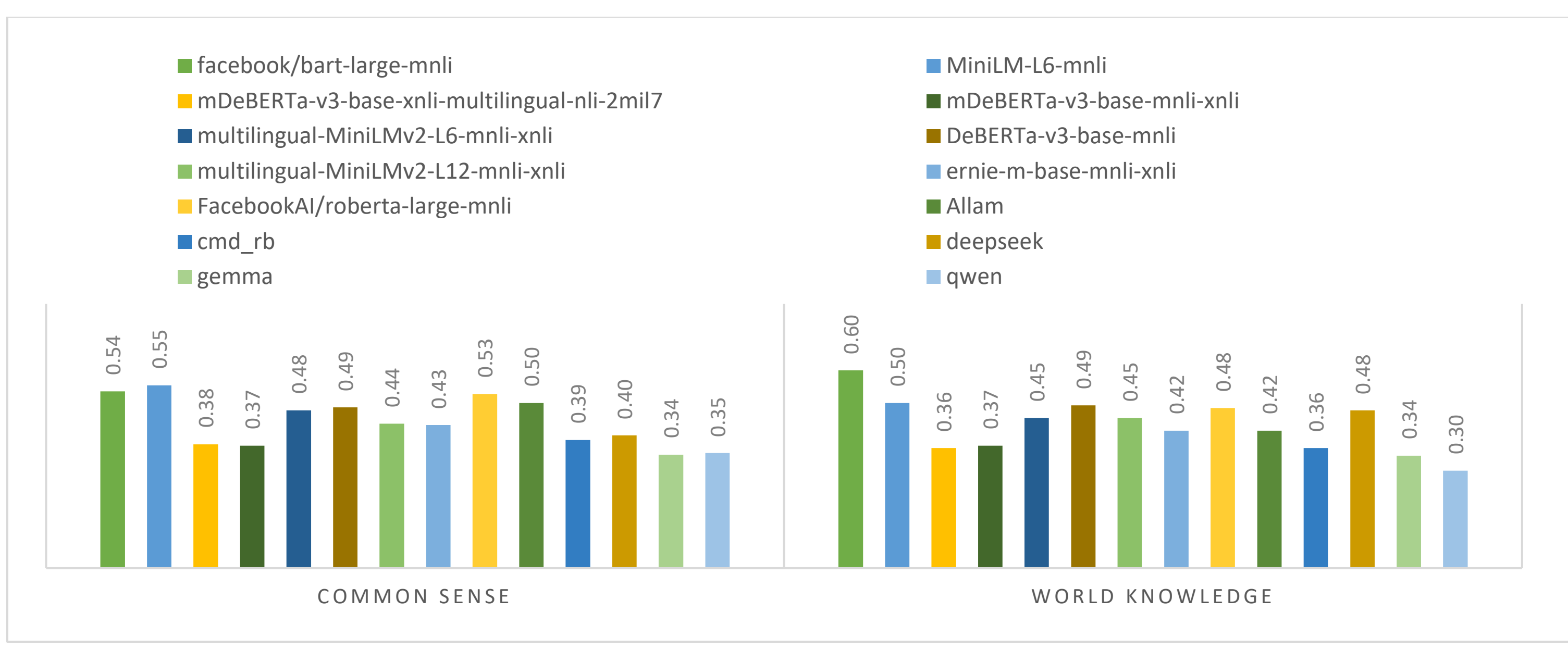

*Figure 7: Common Sense & World Knowledge Fine-grained Error Rates on E-CONAN Diagnostics Data*

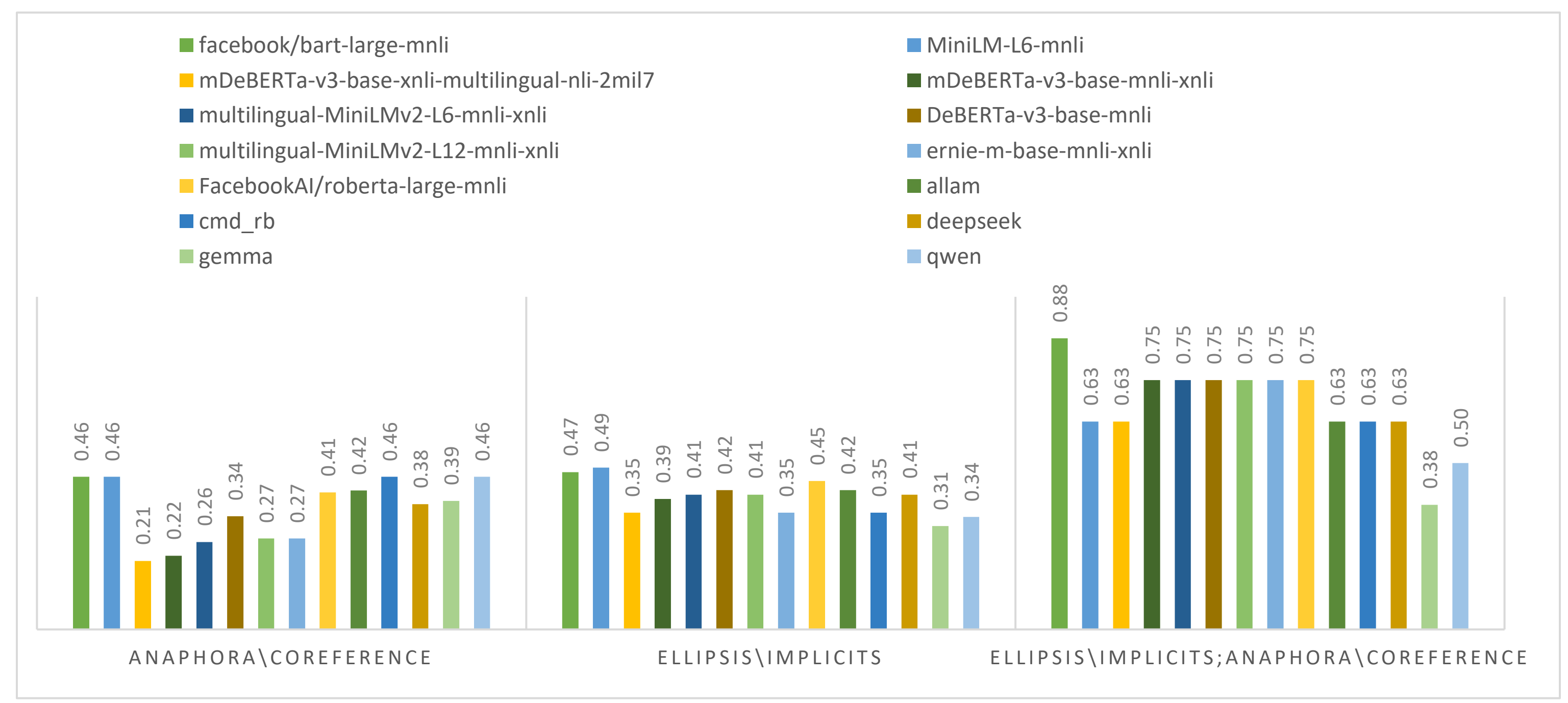

*Figure 8: Discourse Fine-grained Error Rates on E-CONAN Diagnostics Data*

As for Syntactic fine-grained results, Figure 9 shows that the highest error rates were in cases of Adjectivities and Exceptions. This may be caused by the difficulty of associating attachment for the adjectives with right nouns and the particles 'إلا', 'سوى' with right clause of sentence. The lowest error rates were for Condition, Topicalization, Idafa, Dative, and Attention. Moreover, what was obviously noted that Condition has zero errors in all models which is a good sign of having enough training samples in all datasets.

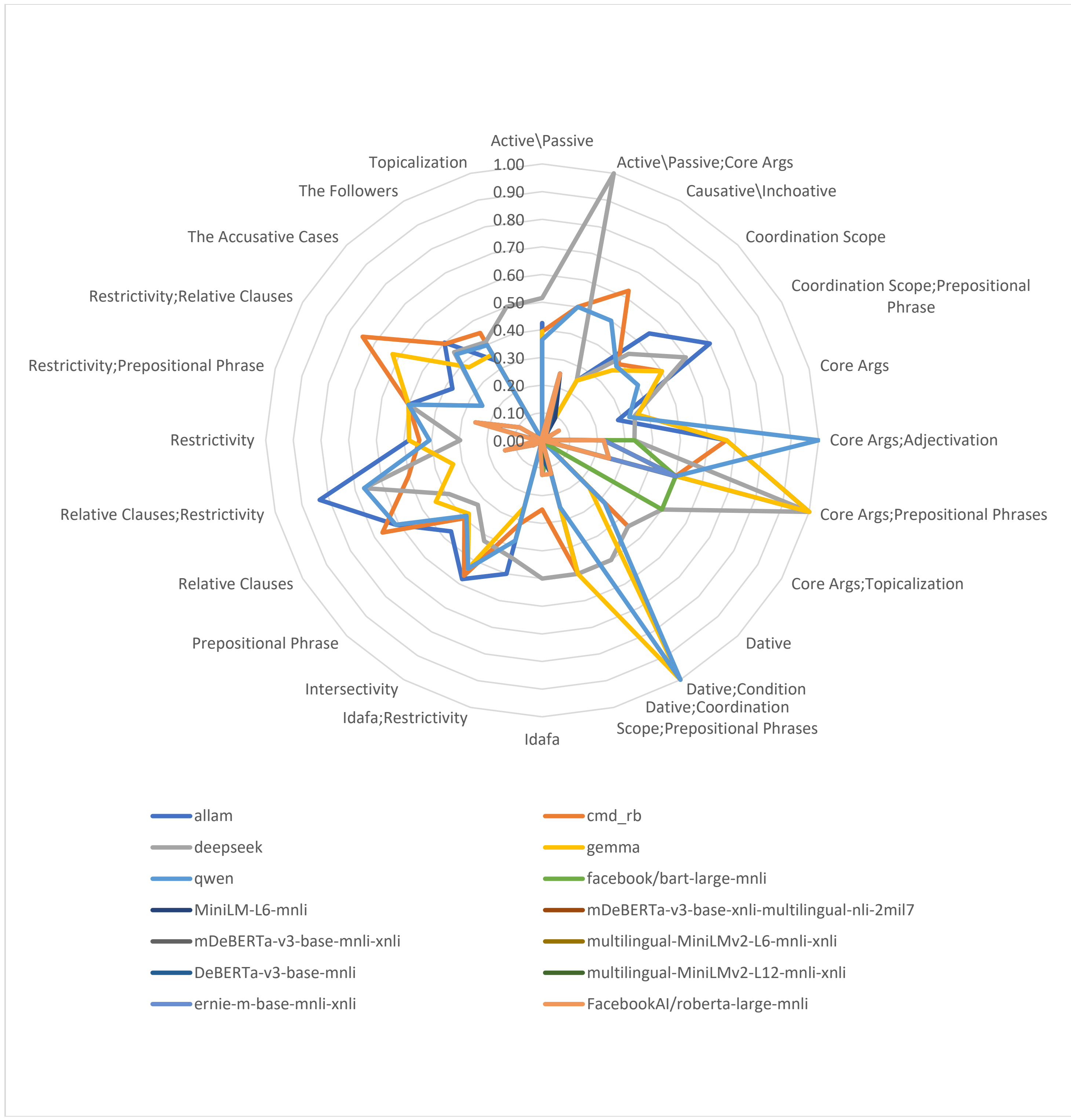

*Figure 9: Syntactic Fine-grained Error Rates on E-CONAN Diagnostics Data*

As for Lexico-Syntactic fine-grained results, Figure 10 show that the lowest error rates were for Nominalization and Genitives\Partitives for all models. Where the highest error rates were for Focus and Interrogatives particles. M4 and M5 have zero error on Affixation and Transparent Head. M1 has zero error rate in Adjectivation, Genitive/Partitives.

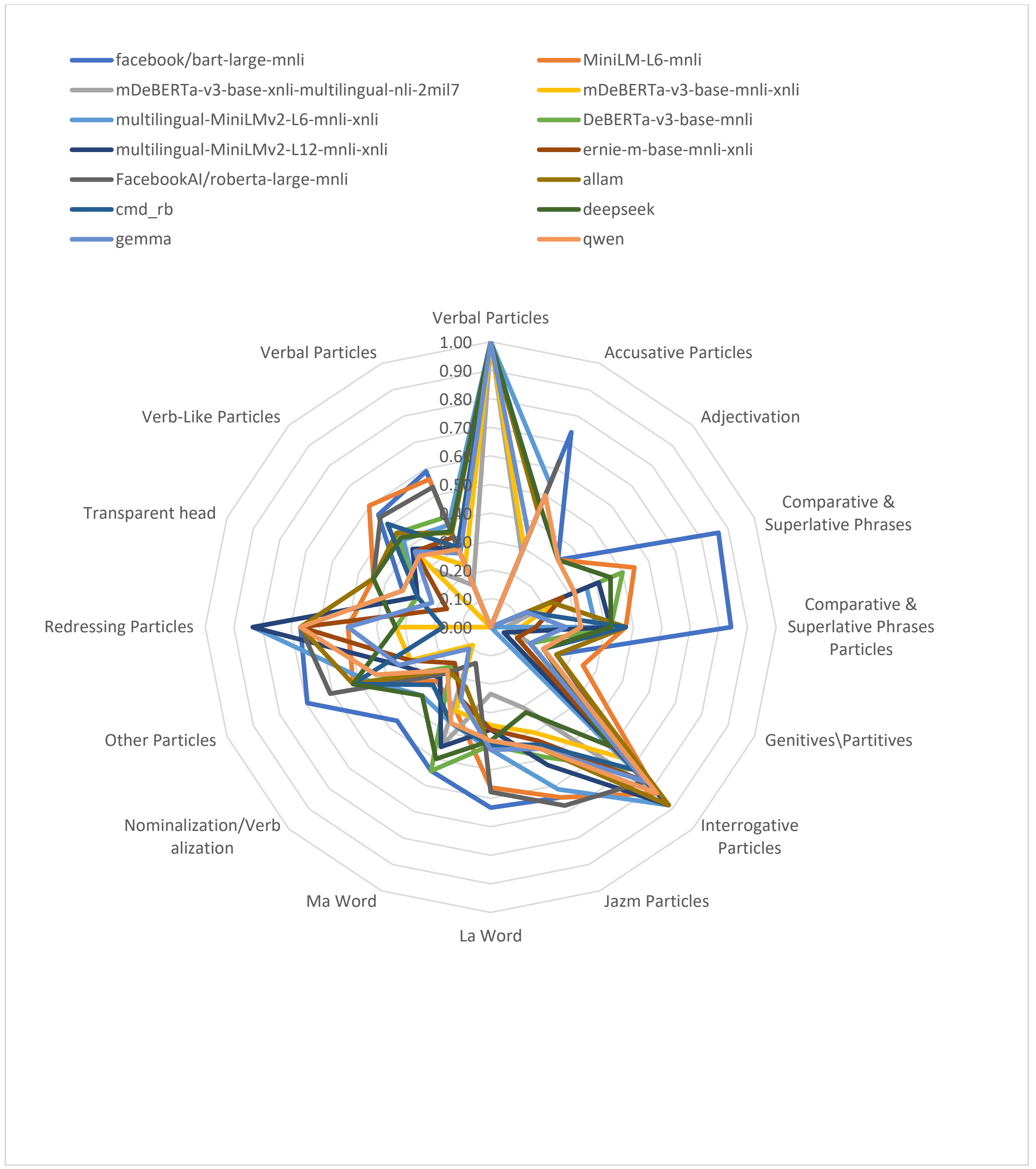

*Figure 10: Lexico-Syntactic Fine-grained Error Rates on E-CONAN Diagnostics Data*

As for Logical fine-grained results, Figure 11 shows that the highest error rates were for the Downward-Monotone and Disjunction. Pretrained models trained on XNLI dataset shows better results in all micro-categories especially in Negation, Double Negations, Conditions and all types of Monotones. LLMs show better or same results as pretrained models.

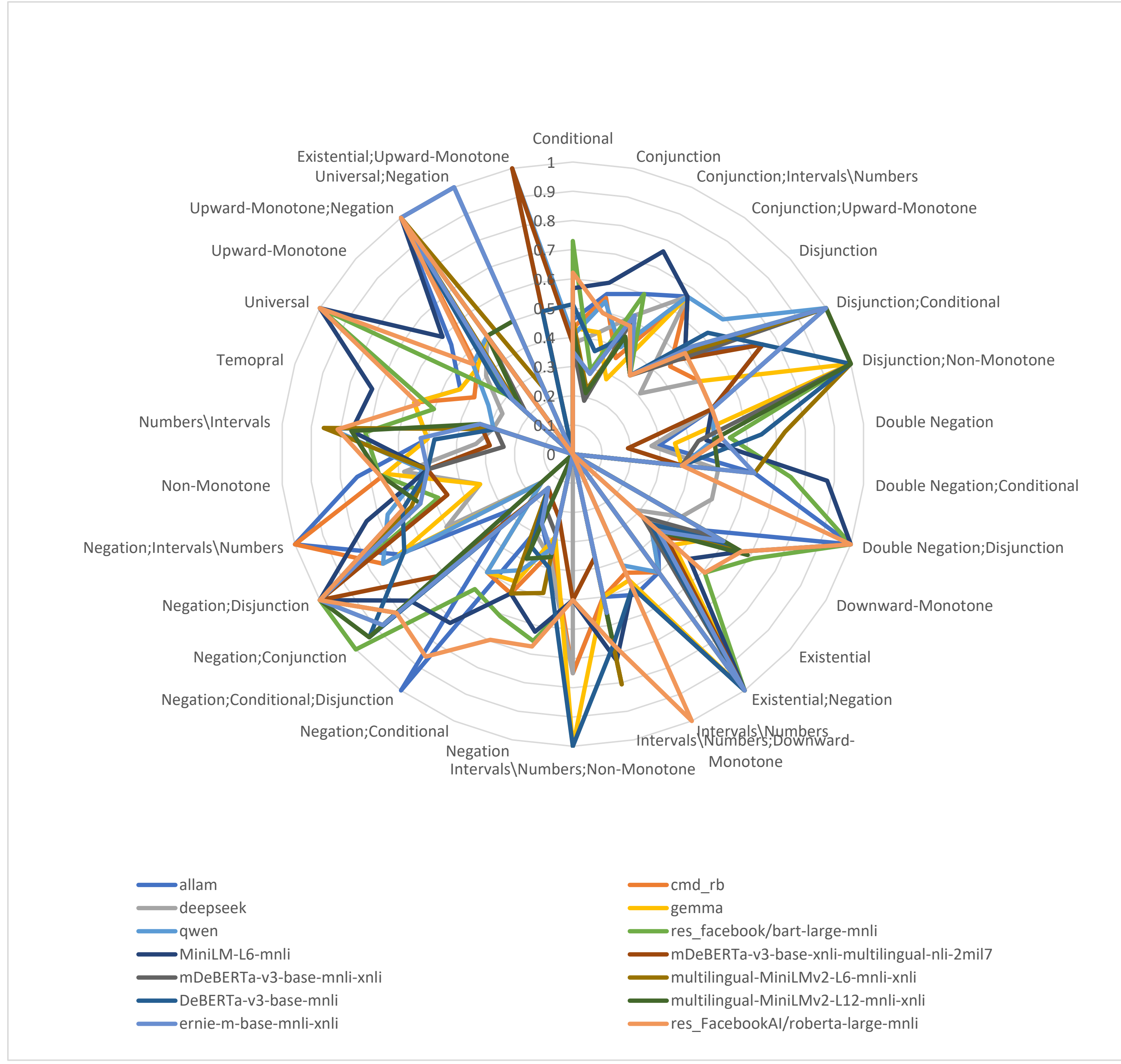

*Figure 11: Logical Fine-grained Error Rates on E-CONAN Diagnostics Data*

## 6. Conclusion

Natural Language Understanding (NLU) is a challenging field due to the complexities of human languages. To improve NLU systems, it's important to analyze errors across different language phenomena. This paper proposed a hierarchical framework for error analysis in Arabic NLU. To implement an example of this framework, we have created a new dataset called E-CONAN Diagnostics. This dataset will be publicly available and includes detailed annotations to help NLU designers understand their models' weaknesses and improve it. We applied E-CONAN Diagnostics error analysis to investigate the performance of 9 pretrained models and 5 LLMs. Results indicate that while the LLMs generally achieved almost equal results with pretrained baselines, LLMs outperform pretrained models in world knowledge and commonsense reasoning macro-category, and underperform pretrained models in syntactic macro-category. Moreover, we note that the hardest phenomena for all pretrained models and LLMs is Reasoning macro-category that contains the highest error rates in all of them. Also, we can note that the easiest phenomena for all pretrained models is Syntactic, and the easiest for LLMs is Lexico-Syntactic. As for LLMs, the highest error rates were by Allam, where the lowest error rate were by Gemma. The highest error rate in all categories was 0.48 and the lowest error rate in all categories was 0.33. As for pretrained models, results show that multilingual pretrained models tuned on both XNLI and MNLI datasets outperform other pretrained models in all coarse-grained categories and in almost all fine-grained categories, and the highest error rate in all categories was 0.63 and the lowest error rate in all categories was 0.24. Although most results were up to 60% error rate, such high error rate indicates the importance of having such difficult benchmarking diagnostics for deeper error investigation.

**ABBREVIATIONS**

**NLI:** Natural Language Inference.
**GLUE**: General Language Understanding Evaluation.
**ALUE**: Arabic Language Understanding Evaluation.
**FRACAS:** Framework for Computational Semantics.

**ACKNOWLEDGEMENT**

We would like to thank our Arabic linguistic experts, Souad Al Jallad, Dr. Abd Al Naser Assaf, Dr. Hanaa Sbinati for their outstanding support and ongoing guidance that was invaluable to the work presented in this paper.

**CONSENT FOR PUBLICATION**

The authors consent for publication.

**AVAILABILITY OF DATA AND MATERIALS**

The datasets will be available online

**COMPETING INTERESTS**

The authors declare that they have no competing interests.

**FUNDING**

The authors declare that they have no funding.

## Appendix

| Macro Category | Level_1 Micro Category | Level_2 Micro Category | Level_3 Micro Category |
|---|---|---|---|
| **Lexical** | Identity/Mismatch | Strings | - |
| | | Numbers | - |
| | | Dates | - |
| | | Format | - |
| | Lexical Semantics | Acronymy | - |
| | | Demonymy | - |
| | | Synonymy | - |
| | | Semantic Opposition (Contrast) | - |
| | | Hypernymy | - |
| | Factivity | Attitudes Verbs | - |
| | | Believe Verbs | - |
| | | Probability Verbs | - |
| | Morphological Negation | - | - |
| | Unique Action Verbs | - | - |
| | Symmetry/Collectivity | - | - |
| | Redundancy | - | - |
| | Named Entities | - | - |
| | Quantifiers | - | - |
| **Lexico-Syntactic** | Genitives/Partitives (Idafa) | Transparent head | - |
| | | With Modifier | - |
| | Alternations | Nominalization/Verbalization | - |
| | | Causative/Inchoative | - |
| | Particles | Waw Letter الواو | - |
| | | Ma Word ما | - |
| | | La Word لا | - |
| | | Annuller Particles[16] (الحروف الناسخة) | Verbal Particles[17] كان وأخواتها |

[16] https://www.madinaharabic.com/arabic-language-course/lessons/L060_001.html

| | | | |
|---|---|---|---|
| | | | Verb-Like Particles[18] إنَّ وأخواتها |
| | | Interrogative Words[19] (أدوات الاستفهام) | - |
| | | Negation Particles[20] (أدوات النفي) | - |
| | | Emphasis Particles[21] (أدوات التوكيد) | - |
| | | The Accusative Particles[22] (أدوات النصب) | - |
| | | The Jazm Particles[23] (أدوات الجزم) | - |
| | | Conditional Particles[24] (أدوات الشرط) | - |
| | | Redressing Particles (أدوات الاستدراك) | - |
| | Comparative Phrases | - | - |
| | Superlative Phrases | - | - |
| **Syntactic** | Argument Realization (Core Arguments) | - | - |
| | The Accusative Cases (المنصوبات) | Absolute Object[25] (مفعول مطلق) | - |
| | | Causal Object[26] (مفعول لأجله) | - |
| | | The Exclusion [27] (مستثنى) | - |
| | | Al-Hal (Status) [28] (الحال) | - |
| | | Al-tamyīz (The Distinctive) [29] (التَّمْيِيز) | - |
| | | Circumstantial Object[30] (مفعول فيه) | - |
| | | Direct Object[31] (مفعول به) | - |
| | The followers[32] (التوابع) | Adjective[33] (النعت) | |
| | | Apposition[34] (Substitution[35]) (البدل) | - |

---

17 https://www.madinaharabic.com/arabic-language-course/lessons/L059_001.html
18 https://www.madinaharabic.com/arabic-language-course/lessons/L060_002.html
19 https://www.madinaharabic.com/arabic-language-course/lessons/L043_001.html
20 https://www.madinaharabic.com/arabic-language-course/lessons/L031_007.html
21 https://www.madinaharabic.com/arabic-language-course/lessons/
22 https://www.madinaharabic.com/arabic-language-course/lessons/
23 https://www.madinaharabic.com/arabic-language-course/lessons/L036_006.html
24 https://www.madinaharabic.com/arabic-language-course/lessons/L049_001.html
25 https://arabikey.com/the-absolute-object-in-arabic/
26 https://www.madinaharabic.com/arabic-language-course/lessons/L068_006.html
27 https://www.madinaharabic.com/arabic-language-course/lessons/L073_001.html
28 https://www.madinaharabic.com/arabic-language-course/lessons/L071_001.html
29 https://www.madinaharabic.com/arabic-language-course/lessons/L072_001.html
30 https://www.madinaharabic.com/arabic-language-course/lessons/L069_001.html
31 https://www.madinaharabic.com/arabic-language-course/lessons/L066_001.html
32 https://www.madinaharabic.com/arabic-language-course/lessons/L078_001.html
33 https://www.madinaharabic.com/arabic-language-course/lessons/L078_002.html

| | | Emphasis[36] (التوكيد) | - |
|---|---|---|---|
| | | Coordinating[37] (العطف) | |
| | Syntactic Semantic Attachment Ambiguity | Sentence Level | Coordination |
| | | | Relative Clauses |
| | | Phrase Level | Prepositional Phrase |
| | | | Adverbial Phrase |
| | Modifier (النعت/الصفة) | Intersectivity (التقاطع) | - |
| | | Restrictivity (التقييد) | - |
| | Alternations | Dative Alternation | - |
| | | Active/Passive Alternation | - |
| | | Topicalization Alternation | - |
| **Discourse** | Anaphora/ Coreference | - | - |
| | Apposition | - | - |
| | Ellipsis/Implicits | - | - |
| **Logic** | Propositional Structure | Conditional | - |
| | | Coordination | Conjunction (عطف توافق) |
| | | | Disjunction (عطف ترافق) |
| | Negation Structure | Negation | - |
| | | Double Negation | - |
| | Quantifiers | Universal | - |
| | | Existential | - |
| | Logical Intervals | Temporal Intervals | - |
| | | Numerical Intervals | - |
| | | Spatial Intervals | - |
| | Monotonicity | Upward Monotone | - |
| | | Downward Monotone | - |
| | | Non-Monotone | - |

[34] https://corpus.quran.com/documentation/apposition.jsp
[35] https://www.fluentarabic.net/ajrumiyyah-arabic-grammar-5/
[36] https://www.madinaharabic.com/arabic-language-course/lessons/
[37] https://www.madinaharabic.com/arabic-language-course/lessons/L079_001.html

| | | | |
|---|---|---|---|
| **Knowledge & Common Sense** | Knowledge | - | - |
| | Common Sense | - | - |
| **Pragmatic** | - | - | - |

*Table 5: Hierarchy of E-CONAN Diagnostics Dataset*

| **Text** | **Hypothesis** | **Class** | **Macro - Micro Categories** | **Explanation** |
|---|---|---|---|---|
| كتب هذا البحث باحثون سوريون | كتب هذا البحث باحثون من سورية | Entail | Lexical-Lexical Semantics-Demonymy | |
| ظننت الامتحان سهلاً | الامتحان سهل | Neutral | Lexical -Factivity-Probability/ Believe Verbs | Imagining an event is not the same as experiencing it. When we use "believe" verbs, we express certainty that the event has occurred. |
| وجدت العلم نوراً | العلم نور | Entail | | |
| ليت الشباب يعود يوماً | الشباب يعود يوماً | Contradict | Lexico-Syntactic-Annuller Particles-Verb-like Particles | "ليت" Particles means hoping something impossible |
| الامتحان سهل | ليت الامتحان سهل | | | |
| المحصول وفير هذا العام | لعلَّ المحصول وفير هذا العام | Contradict | | "لعلَّ" Particles means hoping something may be possible in future, may happen or not. Moreover, we do not hope something already happened. So, we will have contradiction label if the event is in premise as it already happened. |
| لعلَّ النجاح يكون حليفي العام القادم | النجاح سيكون حليفي العام القادم | Neutral | | |
| كسر زيد المزهرية | كسر زيد المزهرية عن غير قصد | Neutral | Lexico-Syntactic-Core Args | While it's ok to have more details and arguments in text, we can't make assumptions in hypothesis that are not clearly provided in text. (Restrictivity) |
| كسر زيد المزهرية عن غير قصد | كسر زيد المزهرية | Entail | | |
| بنى هذا القصر عمرو | بنى هذا القصر زيد | Contradict | Lexical-Unique Action Verbs | "build", "die" are unique actions that happens once for the same entity. While "rebuilt", "travel" are not unique functions and can be repeated over time. |
| رمَّم هذا القصر عمرو | رمَّم هذا القصر زيد | Neutral | | |
| سافر زيد إلى دمشق | سافر زيد إلى حمص | Neutral | | |

| مات زيد في دمشق | مات زيد في حمص | Contradict | | |
|---|---|---|---|---|
| جاء الرجل | جاء الرجل ضاحكاً | Neutral | Syntactic-Accusative Cases-State | It’s the same as restrictivity in logic domain. |
| جاء الرجل ضاحكاً | جاء الرجل | Entail | | |
| أكلت الفاكهة مساءً | أكلت الفاكهة صباحاً | Contradict | Syntactic-Accusative Cases-Circumstantial Object | Different times or places. (ظرف زمان أو ظرف مكان) |
| أكلت الفاكهة في العمل | أكلت الفاكهة في المنزل | Contradict | | |
| أغلق الباب من فضلك | الباب الآن مغلق | Contradict | Pragmatic | We do not request doing something already done. |
| أغلق الباب من فضلك | الباب الآن مفتوح | Entail | | |
| انتشر وباء كورونا في القرن الحالي | انتشر وباء كورونا قبل عدة أعوام | Entail | Pragmatic | |
| انتشر كورونا في القرن الماضي | انتشر وباء كورونا قبل عدة أعوام | Contradict | | |

*Table 6: Examples from E-CONAN Diagnostics Data*